# An AI-Powered VVPAT Counter for Elections in India

**Prasath Murugesan[1], Shamshu Dharwez Saganvali[2]**

[1] PSG College of Technology, Coimbatore

[2] MiQ Digital USA Inc

{prasath.murugesan2001@gmail.com, shamshu@miqdigital.com}

## Abstract

The Election Commission of India has introduced Voter-Verified Paper Audit Trail since 2019. This mechanism has increased voter confidence at the time of casting the votes. However, physical verification of the VVPATs against the party level counts from the EVMs is done only in 5 (randomly selected) machines per constituency. The time required to conduct physical verification becomes a bottleneck in scaling this activity for 100% of machines in all constituencies. We propose an automated counter powered by image processing and machine learning algorithms to speed up the process and address this issue.

## Introduction

The functioning of a healthy democracy is defined by the way elections are conducted in it. Centuries of learning and experimentation have led different countries to adapt to various voting and representation methods to run a democracy. Though there are great technological advancements, the vehicles to collect votes from constituents have not changed from physical ballots in most countries. However, India, the world's largest democracy, has adopted Electronic Voting Machines (EVMs) since 1998 (URL 3). Since then the credibility of the election process has been under severe contention across different entities. Opposition parties have raised doubts about the credibility of electronic voting machines a number of times in the past two years (URL 5). The Election Commission has repeatedly denied allegations that the machines can be tampered with. In spite of this, during every major poll, tens of cases have been filed in the Election commission of India (ECI) questioning the integrity of the EVM machines and the process. One of the predominant claims has been that the voters are unsure about the accuracy of the counter in the machine since they are not sure whether their vote increments the count for the intended party. This was not an issue in the case of paper ballots because the voters would stamp against the party symbol before dropping the ballot paper into the box.

The ECI has been conducting various measures to ensure the authenticity of the EVMs and assures that it is tamper-proof. In 2019, as an additional measure to address this issue and to enhance the trust of the voters, the ECI announced the usage of Voter-Verified Paper Audit Trail (VVPAT) slips as part of the voting process. The whole unit of the EVM machine with the VVPAT machine is shown in Figure 1. When the voter presses the button of the intended symbol (party), the VVPAT slip with the exact symbol will be printed and displayed to the voter through display for about 7 seconds before it gets collected in the bin underneath it. Now the voter has the confidence that their vote has been recorded as per their intention assuming that the counter is tamper-proof.

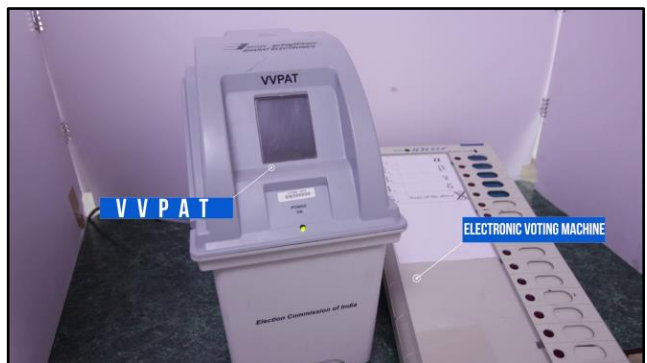

Figure 1: The EVM machine with the VVPAT unit

Before the Lok Sabha Polls in 2019, on the day of counting, the VVPAT slips from one EVM from every assembly segment/constituency (randomly drawn) were subjected to physical verification. The intention is to check whether the EVM's vote tally is accurately verified against the VVPAT slips physically collected (URL 1, URL 2). This is captured in ECI guidelines 16.6 of 2019 (URL 13, URL 14). Later that year, the Supreme Court was hearing a petition filed by 21 opposition parties who sought direction to verify at least 50% of the votes cast in the Lok Sabha elections using the Voter Verified Paper Audit Trail machines. The Supreme Court ordered ECI to increase the number of EVMs subjected to physical verification from 1 to 5 EVMs in every assembly segment/constituency (URL 1, URL 2). This increased the total number of machines that will be subjected to physical verification to 20,625 from 4,125 earlier this year.

## Problem Definition & Proposed Solution

The VVPAT slips have certainly added to the confidence of the voters and the opposition parties. This additional process has, however, brought added difficulties during the counting stage. Two key issues that are identified are the following 1) Mismatch between the EVM and VVPAT still occurs however small the difference be (URL 5, URL 6, URL 7). 2) As the number of EVM machines that are subjected to physical verification increases, the amount of time required to declare the results also increases. In one of the affidavits that the ECI filed in the Supreme court, it cited logistical difficulties in verifying 50% VVPAT slips, which would delay the announcement of results by six days. This would also require extensive training and capacity building of election officials on the field. Without such training, human error will play a vital role in the mismatches. It is to be noted that, as per the 56D (4)(b) of the Conduct of Election Rules, 1961, if there is any discrepancy between the EVM count and VVPAT count the latter prevails, which nullifies the counts from the EVMs (URL 7). In such scenarios, this procedure is in a way equivalent to paper ballot counting. So VVPAT counting is going to be a critical and time-consuming step in all future elections involving EVMs. The authors hereby propose an automated counter which is powered by Image processing and machine learning algorithms to count the VVPAT slips. The automated counter has a hardware layer which will capture the image of each VVPAT slip and will pass the image for classification by a machine learning (ML) layer. A proposed design of a VVPAT counter is provided in Figure 2. The focus of this paper is only concerned with the Image Classification layer. This tool will address both the issues of human error and speed mentioned above along with the scope that the number of EVMs per constituency (subjected to VVPAT verification) can also be increased to 50% or higher to provide greater satisfaction to all contesting parties and the citizenry.

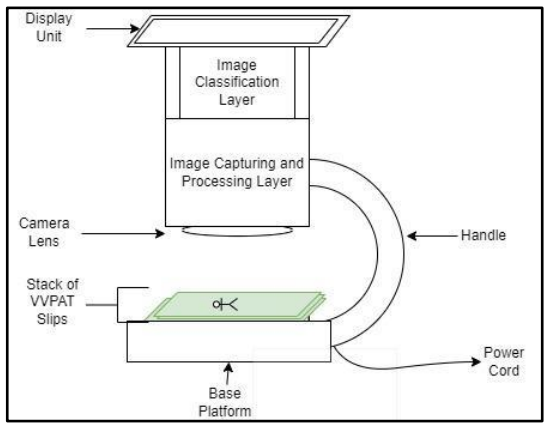

Figure 2: Proposed Design of the VVPAT Counter and its parts

## Data Set

There was no predefined dataset of party symbols to be used as a benchmark. We curated a dataset from various national and regional websites owned by the ECI (URL 4). The dataset consists of symbols (image files) of 49 National and State registered parties approved by the ECI. There will be candidates from newly registered political parties and many independent candidates for whom the ECI will allot symbols only after the nomination process. These new symbols can be plugged into our model and this process will be discussed in the Results and Discussion section. For each image of the original party symbol, 18 different distortions and transformations were created as variations to the training data. An example of the actual party symbol and its variations is provided in Figure 3. Each image is of the dimension 180 x 180. The final labeled dataset consists of 931 images of party symbols with their corresponding party names as the labels. The codes relevant to the generation of images and the final dataset are provided in (URL 9) for reproducibility.

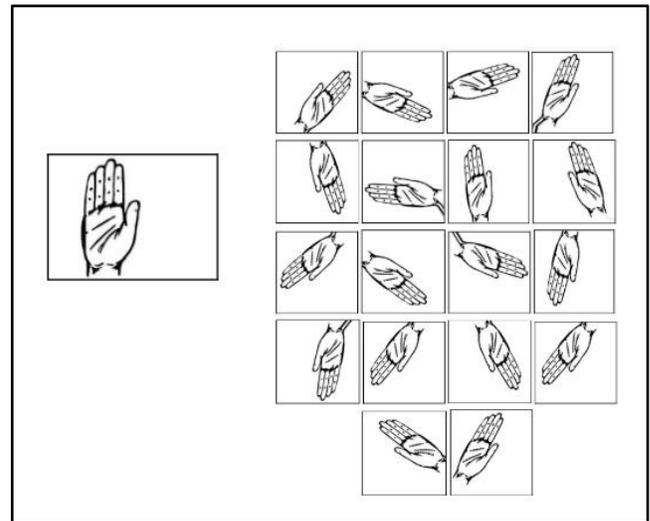

Figure 3: Actual party symbol and its 18 variations

## Experimental Methodology

We have used transfer learning approach (Hussain et. al 2018) approach by leveraging pre-trained CNN models - MobileNetv2 (Sandler et. al 2019) and ResNet50 (He et. al 2015), which are trained on ImageNet dataset (URL 12). We have preserved all hidden layers of the predefined model architecture. We added a custom hidden layer and plugged it into the output layer with 49 neurons corresponding to all 49 parties considered for the experiment with ReLu activation function to learn without fine-tuning pre-trained layers. This approach of transfer learning without fine-tuning saves computation time in the training phase by preserving parameters from all layers except the custom layer.

During the training phase, we used a batch size of 32 training images and trained both models for 15 epochs with a train-test split of 80-20% split. After training the model on our party symbol dataset, we made the model parameters available in our GitHub repository (URL 8, URL 10) as .hdf5 files for reproducibility and can be extended to other models.

## Results and Discussion

We have obtained validation accuracy of 98% and 99% for the MobileNetV2 and ResNet50 models respectively. The training and validation accuracy and their corresponding loss by epoch is given in Figure 4. These plots indicate that the model is not being overfit for the given labeled data and we can expect reliable accuracy during classification in the real-world instances.

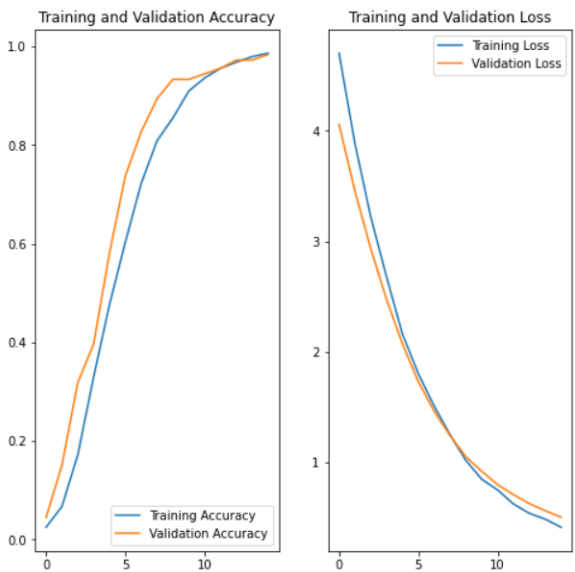

Figure 4: Accuracy and Loss plots for the training and validation sets by Epoch(x-axis) of ResNet50

By analyzing the confusion matrices, we have identified that the recall is less than 0.80 for 6 party symbols that would need additional data samples to improve prediction. So, we recommend any slip below the suggested threshold should be subjected to human verification and the corresponding count has to be added to the respective party at the end of the tally.

We have also identified that some of the national vs state parties have been allocated the same symbols or very similar symbols. For example: An Elephant or A Candle. The base model built is extendable to accommodate learning and classification of new symbols. In such scenarios an additional training will be required with party symbols contesting in that region to avoid mislabeling. Irrespective of the national vs state elections, the models can be trained on the database of symbols allotted at the state level and can be tallied up at the time of counting votes. When both the national party and the state party contest in the same constituencies, the tie will be broken by the ECI allocating unique party symbols to the candidates. So, this scenario should not be a challenge for the model training and implementation.

We served the model in a web app (URL 11) to simulate the vote counting application. The application gets greyscale images from users in JPG and JPEG formats and predicts using the trained model at the backend of the application. In the future extensions of the model we also plan to provide probability scores for the prediction. These probabilities can be used as a threshold and any VVPAT slip that is predicted at lower confidence can be separated by the machine for human intervention.

A polling booth is allotted for every 1500 voters as per the guidelines given by the ECI (URL 15). This means that every EVM will record not more than 1500 votes and the same will apply to the VVPAT slips. So, for each EVM machine, the prediction task will be reduced to labeling (Party symbols) 1500 VVPAT slips. The prediction time for one image is less than 40 milliseconds. If all the VVPAT slips are captured as images and stored, then the total time for predicting 1500 images will be around 1 minute. The time required to capture 1500 images by the prototype discussed in Figure 2 is roughly projected to be 10 minutes. By these estimations, we propose that, for each EVM, the automated labeling of VVPAT slips and aggregating the votes by party symbol would only need 15 minutes or less time. Let us assume a state election with 60 million (6 crores) voters. Such a state would need 40,000 polling booths and the same number of EVM/VVPAT machine pairs for recording the votes. The counting of votes also happens in a distributed fashion i.e, all the EVMs in every district will be brought under one physical location (counting center) and subjected to the counting process. For the purposes of estimation, it is also assumed that we have access to 1500 units of the prototype proposed above and these units are shared appropriately (based on voters' proportion) before the commencement of the counting.

By the estimates mentioned above and assuming we have access to 1500 units of the proposed prototype, 100% of VVPAT slips can be counted and tallied up in 7 hours or less time. Such a scenario will significantly reduce the time required to declare the results than what the ECI has filed in the affidavit (URL 2). This would also satisfy the prayers that the opposition parties filed in the supreme court (URL 2) and the common citizen by leaving no room for question on credibility of election processes since all VVPATs are verified and there will be no unutilized VVPATs. Thus, we conclude that the proposed prototype and the methodology will expedite the counting of VVPATs and will also increase the trust in the election process by a significant degree.

We also make the following recommendation based on our study of VVPAT slips and related issues. The time stamp at which the VVPAT slip was generated can be printed on each slip which will be helpful in identifying voting abnormalities and separating VVPATs from mock polls. In practical voting scenarios, for a given polling unit, no more than 8 votes can be registered in a minute. There is no mechanism today to identify any such anomalies today except the CCTV recordings which are intractable. With this added metadata on the VVPAT slips such anomalous behavior can be easily identified and the corresponding polling officers can be brought in for further investigation by Election Commission of India.

## Conclusion

The introduction of VVPATs by the ECI has increased voter confidence but it brought additional challenges during the counting stage. The purpose of this new system will be achieved only when 100% of VVPATs are subjected to verification against the EVM counts. This would delay the announcement of results by several days according to the current manual process. An automated counter proposed by our work can be a solution to speed up the counting process, so that all EVMs can be subjected to verification and helps to reduce human error to a greater degree. We also believe this work and its recommendations will contribute to the fairness of elections and gain the complete trust of the citizenry.

## References

[Hussain et. al 2018]
Mahbub Hussain, Jordan J. Bird, Diego R. Faria. 2018. A Study on CNN Transfer Learning for Image Classification. UKCI 2018.
[Sandler et. al 2019]
Mark Sandler, Andrew Howard, Menglong Zhu, Andrey Zhmoginov, Liang-Chieh Chen. 2019. MobileNetV2: Inverted Residuals and Linear Bottlenecks. https://arxiv.org/pdf/1801.04381.pdf
[He et. al 2015]
Kaiming He, Xiangyu Zhang, Shaoqing Ren, Jian Sun. 2015. Deep Residual Learning for Image Recognition.

[URL1] https://www.thehindu.com/news/national/increase-random-checking-of-vvpat-slips-sc-tells-election-commission/article26769488.ece
[URL 2]
https://scroll.in/latest/919313/vvpat-verification-supreme-court-orders-counting-of-paper-slips-of-five-evms-in-every-constituency
[URL 3]
https://eci.gov.in/voter/history-of-evm/
[URL 4]
https://en.wikipedia.org/wiki/List_of_political_parties_in_India
[URL 5]
https://www.hindustantimes.com/elections/sc-refuses-urgent-hearing-to-plea-for-enhanced-vvpat-verification-101646773209266.html
[URL 6]
https://thewire.in/rights/evm-miscount-vvpat-citizens-right-secret-verified-ballot
[URL 7]
https://economictimes.indiatimes.com/news/politics-and-nation/eight-cases-of-vvpat-evm-mismatch-in-lok-sabha-polls/articleshow/70323347.cms
[URL 8] https://github.com/Prasath2001/Party-Symbol-Classifier/blob/main/datasetCreation.py
[URL 9]
https://www.kaggle.com/datasets/prasathm2001/indian-party-symbol-dataset
[URL 10]
https://github.com/Prasath2001/Party-Symbol-Classifier
[URL 11]
https://votecounter.streamlit.app/
[URL 12]
https://www.image-net.org/
[URL 13]
https://www.eci.gov.in/files/file/9230-manual-on-electronic-voting-machine-and-vvpat/
[URL 14]
https://pib.gov.in/PressReleasePage.aspx?PRID=1572405
[URL 15]
https://www.thehindu.com/news/cities/chennai/tamil-nadu-will-have-23000-more-polling-booths-says-governor/article33663262.ece